\documentclass[sn-standardnature,iicol]{sn-jnl}

\usepackage{array}
\usepackage{bm}
\usepackage{caption,subcaption}
\usepackage{color}
\usepackage[bottom]{footmisc}
\usepackage{makecell}
\usepackage{makeidx}
\usepackage{mathtools}
\usepackage{multicol}
\usepackage{tabularx}
\usepackage{url}

\def\bfn#1{\bm{\mathbf{#1}}}

\newcommand{\ie}{\textit{i.e.}}
\newcommand{\eg}{\textit{e.g.}}

\jyear{2022}

\begin{document}

\title[Survey on Deep Fuzzy Systems]{Survey on Deep Fuzzy Systems in regression applications: a view on interpretability}

% Authors
\author*[1]{\fnm{Jorge} \sur{S. S. J{\'u}nior}}\email{jorge.silveira@isr.uc.pt}
\author[1]{\fnm{J{\'e}r{\^o}me} \sur{Mendes}}\email{jermendes@isr.uc.pt}
\author[2]{\fnm{Francisco} \sur{Souza}}\email{f.souza@science.ru.nl}
\author[1]{\fnm{Cristiano} \sur{Premebida}}\email{cpremebida@isr.uc.pt}

\affil*[1]{\orgdiv{Department of Electrical and Computer Engineering, P{\'o}lo II}, \orgname{University of Coimbra, Institute of Systems and Robotics}, \orgaddress{\city{Coimbra}, \postcode{3030-290}, \country{Portugal}}}
\affil[2]{\orgdiv{Dept. Analytical Chemistry \& Chemometrics}, \orgname{Radboud University}, \orgaddress{\city{Nijmegen}, \postcode{6525 AJ}, \country{the Netherlands}}}

\abstract{
Regression problems have been more and more embraced by deep learning (DL) techniques. The increasing number of papers recently published in this domain, including surveys and reviews, shows that deep regression has captured the attention of the community due to efficiency and good accuracy in systems with high-dimensional data. However, many DL methodologies have complex structures that are not readily transparent to human users. Accessing the interpretability of these models is an essential factor for addressing problems in sensitive areas such as cyber-security systems, medical, financial surveillance, and industrial processes. Fuzzy logic systems (FLS) are inherently interpretable models, well known in the literature, capable of using nonlinear representations for complex systems through linguistic terms with membership degrees mimicking human thought. Within an atmosphere of explainable artificial intelligence, it is necessary to consider a trade-off between accuracy and interpretability for developing intelligent models. This paper aims to investigate the state-of-the-art on existing methodologies that combine DL and FLS, namely deep fuzzy systems, to address regression problems, configuring a topic that is currently not sufficiently explored in the literature and thus deserves a comprehensive survey.
}

\keywords{Deep Regression, Explainable Artificial Intelligence, Interpretability, Deep Fuzzy Systems}

\maketitle

\section{Introduction}
\label{sec:Introduction}

The goal in regression is to predict one or more variables $\bfn{y}(k)=(y_{1}(k),\ldots,y_{m}(k))^T$ from the information provided by measurements $\bfn{x}(k)=(x_{1}(k),\ldots,x_{p}(k))^T$ for a given sample $k$. Customary, $\bfn{y}(k)$ are refereed as targets, outputs, or dependent variables, while $\bfn{x}(k)$ are commonly refereed as predictors, inputs, covariates, regressors, or independent variables. Regression models covers several application areas, as economic growth problems \cite{Busu2019,Botev2019}, air quality prediction \cite{Liu2021b,Gu2018b}, medicine \cite{Zhu2020,Shi2019b}, chemical industries \cite{Orlandi2021,Yang2021}, and industrial processes \cite{Souza2021,Liu2019}. Recent studies show that regression models have become predominant in increasingly complex real-world systems due to the large availability of data, inclusion of nonlinear parameters, and other aspects intrinsic to the application area. For complex systems, traditional machine learning techniques (\ie, non-deep/shallow techniques) may become limited as they face two main characteristics in current systems: high-dimensionality (with a high amount of observations and inputs) and complexity (due to the variety of dynamic and nonlinear features, hyperparameters, and uncertainty). Deep learning (DL) techniques have gained prominence in recent years due to their ability to represent systems in complex structures with multiple levels of abstraction and high-level features. However, deep learning techniques may have limitations such as the need for a sufficiently large dataset for model training, sensitivity to hyperparameter selection, and interpretability issues \cite{Sun2021,Angelov2020}. In this sense, deep fuzzy systems have emerged as a viable methodology to balance accuracy and interpretability in complex real-world systems.

Regression models can be categorized as \cite{Sjoberg1995}: (i) ``white-box'' when the input-output mapping is built upon first-principle equations, (ii) ``black-box'' when the mapping is derived from the data (also referred to as data-driven modeling), or (iii) ``grey-box'' when the knowledge about the input-output mapping is known beforehand and integrated along with data-driven modeling. White-box models are advantageous for promoting interpretations of the internal mechanisms associated with input-output mapping. On the other hand, black-box models can address complex systems with predictive analysis without prior knowledge of the system. Grey-box models can combine the interpretability presented by ``white-box'' and the ability to learn from data given by ``black-box'' models (\eg, fuzzy systems). Regarding data-driven modeling, the dependency between input and output can be built by linear or nonlinear models. A regression model is linear when the rate of change between input-output is constant due to the linear combination of the inputs. Examples include the multivariate linear regression models. Data-driven nonlinear regression is adopted when the input-output dependence is nonlinear and can not be covered by linear modeling. There is a plethora of methods for nonlinear regression, and its applicability is problem-dependent. Examples include fuzzy systems, support vector regression, artificial neural networks (\eg, non-deep/shallow and deep networks), and rule-based regression (\eg, decision trees and random forest). 

Recent surveys and reviews have showcased the application of DL to regression problems. The work of Han et al. \cite{Han2021} reviews deep models for time-series forecasting, where the models are categorized as: (i) discriminative, where the learning stage is based on the conditional probability of the output/target given an observation, (ii) generative, which learn the joint probability of both output and observation, with the generation of random instances), or (iii) hybrid, a combination of different deep methods. With the implementation in benchmark systems and a real-world use case related to the steel industry, the authors showed that deep models are efficient in discriminating complex patterns in time series with high-dimensional data. Sun et al. \cite{Sun2021} discuss the use of DL for soft sensor applications, showing the trends and applications in industrial processes, in addition to best practices for model development. The authors established some directions for future research, such as solutions to address the lack of labeled samples (\eg, semi-supervised methods), hyperparameter optimization, solutions to improve model reliability (\eg, model visualization), and the development of DL methods with distributed and parallel modeling. Torres et al. \cite{Torres2021} explore deep learning for time-series forecasting, together with time-series definition and processing using deep architectures commonly described in the literature. In addition, the authors demonstrated some practical aspects of using DL methods to solve complex problems with big data, such as a variety of libraries and techniques compatible with deep structures for automatic optimization of hyperparameters, hardware infrastructure to benefit DL implementation with optimizable complexity and processing time, and flexibility to address real-world applications. Pang et al. \cite{Pang2021} and Chalapathy et al. \cite{Chalapathy2019} present an overview of DL studies for anomaly detection; they also discuss the complexities and different types of models (\eg,\ classification, autoregressive, unsupervised, and semi-supervised) applied in various intelligent systems, such as cyber-security systems, medical monitoring, financial surveillance, and industrial processes. Other works that review the DL literature on regression are \cite{Dong2021,Sengupta2020,Pouyanfar2018}. It is noted from these works that DL techniques have some advantages over non-deep methods, such as the ability to learn complex representations with automatic feature engineering, not requiring prior experience or knowledge, good performance by increasing the dimensionality of the data, among other specific advantages depending on the type of framework and application \cite{Navamani2019}.
 
Despite the advantages portrayed in the literature, deep learning has limitations, such as the need for a sufficiently large dataset for model training, sensitivity to hyperparameter selection, and lack of interpretability \cite{Sun2021,Angelov2020}. Also, there is a lack of proper explanations of the internal structure of deep model structures, which raises concern in applications that directly and indirectly impact human life as well for operational decisions  \cite{Angelov2021b,Confalonieri2021}. Having this concern in mind, recent works address the interpretability issue of DL from the following principles of ``eXplainable Artificial Intelligence'' (XAI) systems \cite{Phillips2021}: the existence of an appropriate explanation for each decision made; each explanation must be meaningful to the user; the process must accurately consider what happens in the system; identify the situations in which the system may or may not function properly (knowledge limits).

Fuzzy logic systems (FLS), composed of IF-THEN rules with linguistic terms mimicking human thought \cite{Chimatapu2018}, is one of the research areas that contemplates the XAI principles. FLS has a wide range of approaches to nonlinear systems, primarily in terms of interpretability, whether related to the complexity and semantics of fuzzy rules, notation readability, coverage of input data space (operation regions), and so on \cite{Lapa2017,as2020}. Moral et al. \cite{Moral2021} show the main benefits of adopting FLS for the development of explainable methodologies, with a discussion of fundamental concepts and definitions associated with XAI and FLS to how to design increasingly interpretable models. Therefore, DL techniques can be complemented with FLS, developing deep fuzzy systems (DFS) and then providing an easy-to-understand and easy-to-implement interface to efficiently address the main drawbacks of DL and FLS, thus ensuring good accuracy and good interpretability. The surveys of \cite{Das2020} and \cite{Zheng2021} investigated some recent trends in DFS models and real-world applications (\eg,\ time series forecasting, natural language processing, traffic control, and automatic control). However, despite the benefit of adopting FLS principles to DL systems, no comprehensive survey or review has been conducted focusing exclusively on deep fuzzy for regression problems.

This paper surveys and discusses the state-of-the-art on deep fuzzy techniques developed to deal with a diverse range of regression applications. Initially, Section \ref{sec.back} presents fundamental concepts about FLS and XAI. Then, an overview of DL techniques commonly used for regression will be presented in Section \ref{sec.DeepRegression}, namely Convolutional Neural Networks, Deep Belief Networks, Multilayer Autoencoders, and Recurrent Neural Networks. Next, Section \ref{sec.DeepFuzzyReg} shows the literature on deep fuzzy systems in two ways: (i) standard deep fuzzy systems, based on fundamental FLS; (ii) hybrid deep fuzzy systems, with the combination of FLS and the conventional deep models discussed in Section \ref{sec.DeepRegression}. Finally, Section \ref{sec.Conclusions} presents general discussions based on the state-of-the-art surveyed.

\section{Background}
\label{sec.back}

\subsection{Fuzzy logic systems}
\label{subsec.OverviewFLS}

Developed initially by Lotfi A.\ Zadeh \cite{Zadeh1965}, FLS are rule-based systems composed mainly of an antecedent part, characterized by an ``IF'' statement, and a consequent part, characterized by a ``THEN'' statement, allowing the transformation of a human knowledge base into mathematical formulations, thus introducing the concept of linguistic terms related to the membership degree. 
  
\begin{figure}[!t]
	\centering
	\includegraphics[width=1\columnwidth]{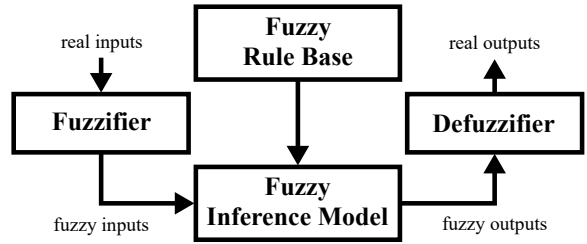}
	\caption{Basic configuration of fuzzy logic systems.}
	\label{fig.basicfuzzy}
\end{figure}

The basic configuration of a fuzzy logic system, shown in Figure \ref{fig.basicfuzzy}, 
depends on an interface that transforms the real input variables into fuzzy sets (fuzzifier), which are interpreted by a fuzzy inference model to perform an input-output mapping based on fuzzy rules. Thus, the mapped fuzzy outputs go through an interface that transforms them into real output variables (defuzzifier) \cite{Wang1997,ci2011}. Some well-known fuzzy systems are Mamdani fuzzy systems \cite{Mamdani1975}, Takagi-Sugeno (T-S) fuzzy systems \cite{Takagi1985}, and Angelov-Yager's (AnYa) fuzzy rule-based systems using data clouds \cite{Angelov2011}. Among these, the T-S fuzzy models stand out for their ability to decompose nonlinear systems into a set of linear local models smoothly connected by fuzzy membership functions \cite{Qiu2016}. T-S fuzzy models are universal approximators capable of approximating any continuous nonlinear system, that can be described by the following fuzzy rules \cite{Ying1997}:
\begin{eqnarray} 
    R^{i}:\textbf{IF} ~
    x_1(k)~\text{is}~F_{1}^{i}~\text{and}~\ldots~\text{and}~x_p(k)~\text{is}~F_{p}^{i}
    \nonumber \\
    \textbf{THEN} ~
    \bfn{y}^i(k)=\bfn{f}^{\,i}(x_{1}(k),\ldots,x_{p}(k)),
\label{eq.generic_rule}
\end{eqnarray}
where $R^{i}$ ($i=1,\ldots,N$) represents the $i$-th fuzzy rule, $N$ is the number of rules, $x_1(k), \ldots,x_p(k)$ are the input variables of the T-S fuzzy system, $F^{i}_{j}$ are the linguistic terms characterized by fuzzy membership functions $\mu^{i}_{F^{i}_{j}}$, and $\bfn{f}(x_{1}(k),\ldots,x_{p}(k))$ represents the function model of the system of the $i$-th fuzzy rule \cite{Junior2021}.

\subsection{Explainable artificial intelligence}
\label{subsec.XAI_back}

Although the explainable artificial intelligence (XAI) concept is often associated with a homonymous program formulated by a group of researchers from the Defense Advanced Research Projects Agency (DARPA) \cite{Hall2019}, the principles related to explainability gained strength from the 1970s onwards. The earliest works presented rule-based structures and decision trees with human-oriented explanations, such as the MYCIN system proposed in \cite{Shortliffe1973} developed for infectious disease therapy consultation, the tutoring program GUIDON proposed in \cite{Clancey1979} based on natural language studies, among numerous other systems \cite{Weiss1978,Suwa1982,Swartout1983,Swartout1985}. Although many authors commonly use the terms ``explainability'' and ``interpretability'' as synonyms, Rudin \cite{Rudin2019} discusses the problem of using purely explainable methods to only provide explanations from the results obtained in black-box models (post-hoc analysis), demystifying the importance of developing inherently interpretable methodologies with causality relationships that are understandable to human users.

\section{Deep Regression Overview}
\label{sec.DeepRegression}

Deep neural networks (DNN) have emerged due to their architecture with multiple levels of representation and their remarkable performance in a variety of tasks \cite{Bengio2013}. This section aims to discuss DL techniques commonly employed in regression problems, providing a better understanding and context for deep fuzzy regression in Section \ref{sec.DeepFuzzyReg}. 

\subsection{Convolutional Neural Networks}
\label{subsec.CNN}

\begin{figure*}[!t]
	\centering
	\includegraphics[width=0.8\textwidth]{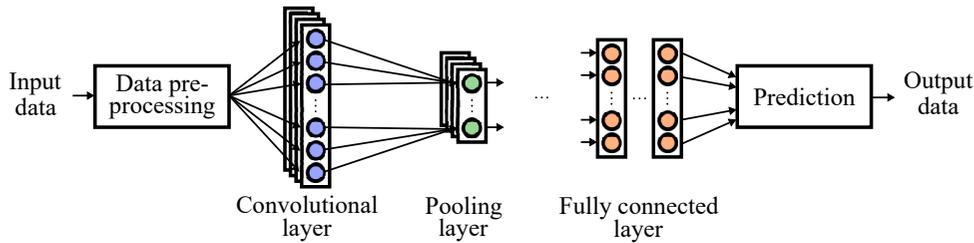}
	\caption{Convolutional neural network, with 1-D architecture.}
	\label{fig.cnn}
\end{figure*}

Convolutional neural networks (CNNs or ConvNets) are feedforward neural networks with a grid-like topology that are used for applications such as time-series data processing in 1-D grids and image data processing in 2-D pixel grids \cite{Goodfellow2016}. Figure \ref{fig.cnn} 
\relax
presents a CNN architecture for processing time-series data in 1-D grids. The ``neocognitron'' model proposed in \cite{Fukushima1980} is frequently referred to as the inspiration model for what is currently known about CNNs. First proposed in \cite{Hubel1968}, the neocognitron aimed to represent simple and complex cells from the visual cortex of animals which present a mechanism capable of detecting light in receptive fields \cite{Gu2018}. 

CNNs use a math operation on two functions called ``convolution'', with the first function referred to as input, the second function as kernel, and the convolution's output as the feature map \cite{Goodfellow2016}. Outputs from the convolution layer go through a pooling layer (downsampling), which performs an overall statistic of the adjacent outputs by reducing the size of the data and associated parameters via weight sharing \cite{Sun2021,Li2021}. After the data is processed along with the layers that alternate between convolution and pooling, the final feature maps go through a fully connected (dense) layer to extract high-level features. For regression problems, the extracted features can be combined in a prediction mechanism with an activation function or a supervised learning model (\eg,\ support vector regression) to estimate the final output \cite{Ketkar2017,Mallat2016}.

The application of CNNs for regression has been explored for traffic flow forecasting \cite{Wu2022,Zhang2020c}, prediction of natural environmental factors \cite{Mukhtar2022,Heo2021,Liu2021c,Wan2019}, industrial process optimization \cite{Gao2021,FAN2020,Yuan2020d}, electrical/power systems applications \cite{Jalali2021,Eskandari2021,Zahid2019,Koprinska2018,Tian2018}, and chemical process analysis \cite{Gao2020,Wu2017}. Despite showing good performance with the extraction of spatially organized features, essentially for pre-processing, CNN's performance depends on a large amount of data and the correct choice of hyperparameters, being computationally intensive \cite{Navamani2019,Witten2017}.

\subsection{Deep Belief Networks}
\label{subsec.DBN}

\begin{figure}[!t]
	\centering
	\includegraphics[width=1\columnwidth]{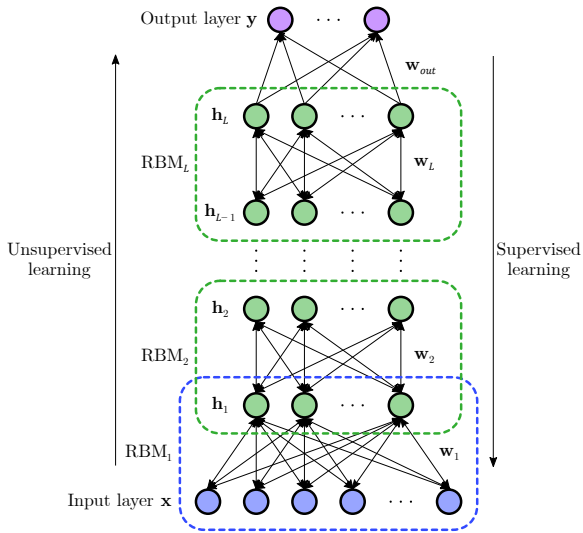}
	\caption{Deep Belief Networks architecture.}
	\label{fig.dbn}
\end{figure}

Deep Belief Networks (DBNs) are probabilistic generative models proposed by Hinton et al. \cite{Hinton2006}, which have a hierarchical structure as illustrated in Figure \ref{fig.dbn}.
\relax
The DBNs are composed of multiple layers of latent variables (hidden units) of binary values, organized into multiple learning modules called restricted Boltzmann machines (RBMs). 

Each RBM comprises a layer of visible units for data representation and a layer of hidden units for feature representation, learned by capturing higher-order correlations from the data. The two RBM layers, with no connections within layers, are connected by a matrix of symmetrically weighted connections, with a total of $L$ weight matrices $\bfn{W} = \{\bfn{w}_{1},\bfn{w}_{2},\ldots,\bfn{w}_{L}\}$, considering a DBN of $L$ hidden layers. All units in each layer have a bidirectional connection with all units in neighboring layers, except for the last two layers, $L$ and output, that have a unidirectional connection \cite{Goodfellow2016}. In the DBN architecture of Figure \ref{fig.dbn}, the layer of visible units in the first RBM represents the input variables $\bfn{x}$ and the subsequent layers represent hidden units $\bfn{h}$, progressing hierarchically until reaching the estimated output, with the output weights $\bfn{w}_{out}$.

Recent studies with DBNs in the context of regression were mainly applied in industrial processes for process monitoring \cite{Wang2021b,Hao2021,Yuan2021,Yuan2020c,Hao2020,Zhu2019}, soft sensing \cite{Wang2021c,Lian2020}, and prognostics \cite{Tian2020,Shao2018}. Other applications include time-series forecasting \cite{Tian2022,Xu2019,Xie2018,Li2017,Jia2016} and benchmark systems modeling \cite{Qiao2018,Qiu2014}. Studies with DBNs have difficulties in investigating the influence of hidden units on system dynamics, leading to interpretability issues. As Figure \ref{fig.dbn} shows, a DBN needs to undergo unsupervised and supervised learning, which allows the training process to become slower as the DBN structure increases. In this way, the DBN becomes sensitive to noisy inputs by not correctly readjusting its low-level parameters \cite{Salakhutdinov2010}.

\subsection{Multilayer Autoencoders}
\label{subsec.SAE}

\begin{figure}[!t]
	\centering
	\includegraphics[width=1\columnwidth]{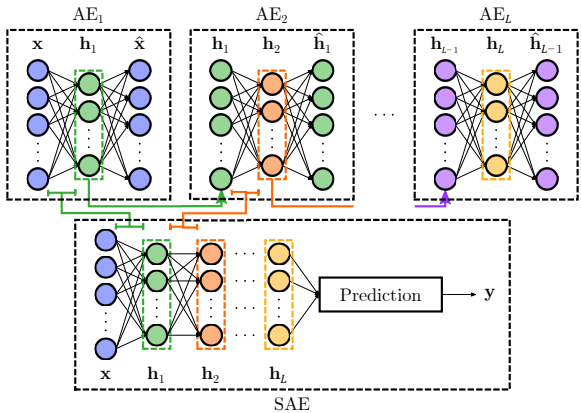}
	\caption{Modified Stacked Autoencoder architecture for regression.}
	\label{fig.sae}
\end{figure}

Autoencoders (AEs) are feedforward neural networks used for dimensionality reduction and representation learning, whose training is aimed at mimicking inputs to outputs \cite{Goodfellow2016}. A historical overview of deep learning in \cite{Schmidhuber2015} presented some early works of AEs in the literature, such as the work in \cite{Ballard1987} that proposes unsupervised architectures to reconstruct the inputs through internal representations. Another early work was published in \cite{Rumelhart1985}, where the authors explore the effect of hidden units in simple two-layer associative networks, in which they want to map input patterns to a set of output patterns. As presented in Figure \ref{fig.sae}, a single AE essentially has a structure composed of a layer with input variables $\bfn{x}$, a layer with hidden units $\bfn{h}$ (which performs an encoding used to represent the input), and an output layer with the reconstructed inputs represented with a ``hat'' symbol (\eg, $\hat{\bfn{x}}$). These layers are interconnected by an encoder function (between input and hidden) and by a function decoder (between hidden and output) \cite{Goodfellow2016}. As an evident constraint, the number of neurons in the input layer must be the same number of neurons in the output layer, setting AE to unsupervised pre-training or feature extraction \cite{Baldi2012}. 

A viable alternative to deal with increasingly complex data is to increase the number of hidden layers in standard AEs, enabling the development of deep network architectures. A well-known configuration of multilayer autoencoders found in the literature is stacked autoencoder (SAE). As illustrated in Figure \ref{fig.sae}, an SAE is developed from the grouping of $L$ AEs, where the hidden layers of the autoencoders are stacked hierarchically, performing an unsupervised layerwise learning algorithm. Thus, the reconstruction of inputs after the $L$-th hidden layer can be disregarded to address regression problems. As for CNNs, a prediction mechanism or a supervised learning model can be included after the $L$-th hidden layer to estimate the system output. Related parameters, such as weights $\bfn{W} = \{\bfn{w}_{1},\bfn{w}_{2},\ldots,\bfn{w}_{l}\}$ between layers, are fine-tuned by a supervised method (\eg, backpropagation algorithm) \cite{Liu2018}.

Recent studies involving types of AEs with deep architecture for regression applied to various cases of industrial processes, such as soft sensing \cite{Sun2021b,Sun2020b,Wang2018b}, hydrocracking process \cite{Liu2021,Yuan2020b,Wang2020c}, CNC turning machine \cite{Shi2019,Bose2018}, end-point quality prediction \cite{Zhang2020,Liu2020}, and prognostics \cite{Wei2022,Li2019,Ren2018}. Other works were developed for time-series forecasting \cite{Li2022,Jin2022,Xiao2021,Lv2018,Jiao2018}. Some of the limitations of multilayer autoencoders include the sensitivities to errors or loss of information from the first layer, impairing learning as it progresses through the hidden layers. With this, the nature of encoding and decoding by hidden layers can cause a loss of interpretability and an increase in computational cost \cite{Navamani2019}.

\subsection{Recurrent Neural Networks}
\label{subsec.RNN}

\begin{figure}[!t]
	\centering
	\includegraphics[width=1\columnwidth]{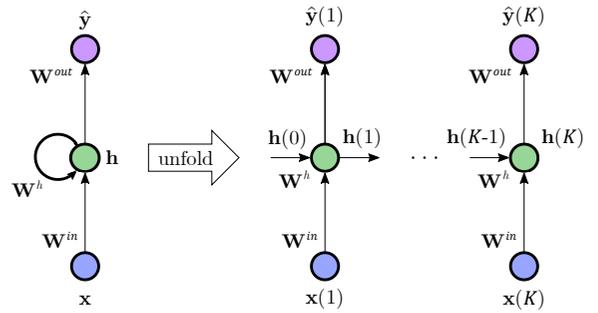}
	\caption{Recurrent Neural Network architecture.}
	\label{fig.rnn}
\end{figure}

Recurrent Neural Networks (RNNs) are artificial neural networks with an internal state that allow the use of feedback signals between neurons \cite{Grossberg2013}. One of the early works that culminated in the popularization of RNNs was proposed in \cite{Hopfield1982}, with the development of content-addressable memory systems called Hopfield networks whose dynamics have a Lyapunov function (or energy function) to direct to a local minimum of ``energy'' associated with the system states \cite{Hopfield2007}. Another early work, published in \cite{Rumelhart1986}, presented a new learning procedure, backpropagation, which adjusts the weights of connections of recurrent networks with ``internal hidden units''. Due to the presence of an ``internal memory'', RNNs are often used to process data in the time domain (sequential information), with weight sharing through hidden states \cite{Zhang2019}. 
Figure \ref{fig.rnn} 
\relax
shows the architecture of an RNN, whose hidden states $\bfn{h}$ represent the ``internal memory'' of the system, and as the inputs are sequentially observed, the corresponding outputs are estimated. The weights $\bfn{W}^{in}$, $\bfn{W}^{h}$ and $\bfn{W}^{out}$ are auxiliary parameters that are shared across time \cite{Albertini1998}. The states are updated at every temporal instant until completing all the input sequences of length $K$ \cite{Pascanu2013,Theodoridis2020}. The RNN learning process depends on a ``backpropagation through time'' algorithm, which is a gradient-based technique that updates parameters recursively starting from the last temporal instant and going backward in time \cite{Werbos1990}.

In deep learning, there are many variations of standard RNNs, such as Long Short-Term Memory networks (LSTMs), Gated Recurrent Units (GRUs), and Echo State Network (ESNs) \cite{Goodfellow2016}. Some of these variations were proposed to cope with some limitations present in standard RNNs, such as exploding and vanishing gradients (instability in networks caused by a large variation in model parameters), overfitting, and difficulty to store low-level features in long data sequences \cite{Lalapura2021}. In this document, only LSTMs and ESNs employed for regression problems will be discussed.

\subsubsection{Long Short-Term Memory}
\label{subsubsec.LSTM}

\begin{figure}[!t]
	\centering
	\includegraphics[width=1\columnwidth]{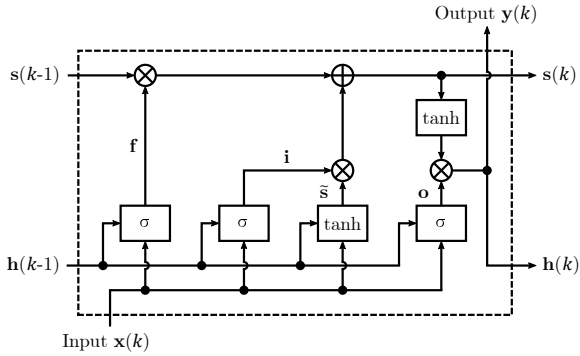}
	\caption{The Long Short-Term Memory cell.}
	\label{fig.lstm}
\end{figure}

Long Short-Term Memory networks (LSTMs) were initially developed in \cite{Hochreiter1997} to overcome the issues of RNNs associated with vanishing/exploding gradients. These issues can occur during the training of long temporal sequences with the backpropagation through time algorithm. In this sense, during successive operations in compound functions with the weight matrices, gradients can exponentially reach very low values close to zero (vanishing) or very high values (exploding) \cite{Goodfellow2016}. Figure \ref{fig.lstm} illustrates the architecture of an LSTM.

LSTMs introduced ``gates'', nonlinear elements that control memory cells using sigmoidal functions $\sigma$, hyperbolic tangent functions, current observation $\bfn{x}(k)$ and hidden units $\bfn{h}(k-1)$ from the previous time instant \cite{Theodoridis2020}. Each of these memory cells has input, output and forget gates ($\bfn{i}$, $\bfn{o}$ and $\bfn{f}$, respectively), that protect the information from perturbations caused by irrelevant inputs and irrelevant memory contents \cite{Witten2017}. The information stored in the cells represents the states $\bfn{s}(k)$ obtained from the data processed in the time domain. Despite having the same inputs and outputs as a standard RNN, an LSTM cell has an internal recurrence (self-loop) to propagate the information flow through a long sequence and, therefore, more parameters to be adjusted \cite{Goodfellow2016}.

Due to the ability to process data sequentially, LSTMs have been used mainly in applications involving time-series forecasting, such as traffic flow \cite{Lu2021,Yang2019} and natural environment factors \cite{Roy2022,Kumari2021,Altan2021,Ma2020,Shen2020,Zhang2017}. Other recent LSTM methodologies for regression have been applied in industrial processes for soft sensing \cite{Zhu2021,Yuan2021b,Yuan2020}, process monitoring \cite{Cai2020,Cheng2019,Li2018} and prognostics \cite{Cheng2022,Guo2021}, as well as applications involving electrical/power systems \cite{Moradzadeh2020,Narayan2017}. Despite the practicality of LSTMs in regression problems, they can still suffer from vanishing due to the possibility of saturation of cell states, which must be reset occasionally, in addition to requiring a high memory bandwidth depending on the execution of functions inside the cell in complex systems (computational cost issue) \cite{Gers2001}.

\subsubsection{Echo State Network}
\label{subsubsec.ESN}

\begin{figure}[!t]
	\centering
	\includegraphics[width=1\columnwidth]{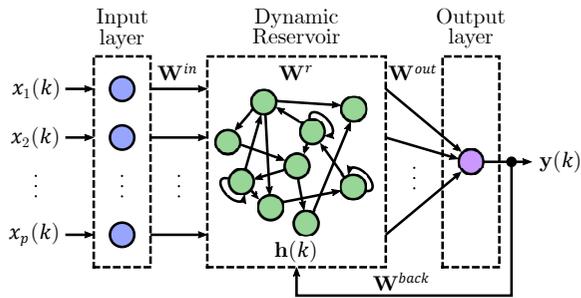}
	\caption{The Echo State Network.}
	\label{fig.esn}
\end{figure}

Echo State Networks (ESNs) are variations of RNNs that were developed in \cite{Jaeger2001} and share the basic ideas of reservoir computing of Liquid State Machines from \cite{Maass2002}. The term ``reservoir computing'' stands for a homonymous research stream that introduced the concept of a dynamic reservoir, in place of the hidden layer, with many sparsely connected neurons \cite{Schrauwen2007}. In addition, this reservoir must have a stability condition known as ``echo state property'', which allows for a gradual reduction in the effect of previous states and inputs on future states over time \cite{Jaeger2009}. Figure \ref{fig.esn} illustrates the architecture of an ESN.

An ESN induces nonlinear response signals from the input signals $\bfn{x}(k)$, whose resulting state $\bfn{h}(k)$ echoes the input information, estimating a desired output signal $\bfn{y}(k)$ \cite{Jaeger2001}. In addition to the input weight $\bfn{W}^{in}$ and output weight $\bfn{W}^{out}$ that are present in a standard RNN, the ESN features the reservoir weights $\bfn{W}^{r}$ and, optionally, the output-to-reservoir feedback weights $\bfn{W}^{back}$. The fundamental idea for the functioning of an ESN is to adjust only $\bfn{W}^{out}$ during training, while the rest are randomly assigned and fixed before learning \cite{Jaeger2007}.

A recurring application of ESN is the time-series prediction in chaotic systems \cite{Tang2022,Na2021,Yao2019,Xu2019b,McDermott2018}, wind power systems \cite{Lian2022,Bai2021,Hu2021,Tian2021,Lopez2018}, solar energy systems \cite{Zhang2021,Li2020,Xu2018,Sun2017}, and benchmark systems \cite{Li2022b,Wang2022,Gao2021b}. Other applications include industrial processes \cite{Schwedersky2022,He2020,Huang2020}, electrical/power systems \cite{Mansoor2021,Yao2019b,Wen2019}, and turbofan engine time-series prediction \cite{Bala2020,Zhong2017}. It is observed in these studies that the random connectivity within the dynamic reservoir of an ESN can lead to interpretability issues. In addition, ESN hyperparameters, such as the number of units in the reservoir and input scaling, have a short operating space that ensures maximum model performance \cite{Xu2020}.

\section{Deep Fuzzy Regression}
\label{sec.DeepFuzzyReg}

Deep fuzzy systems (DFSs) are models built on top of DL structures with fuzzy logic systems (FLSs). DFS aims to overcome the lack of interpretability of DL systems and the limitations of FLS when dealing with high-dimensional data. DFS is often referred to in the literature as an explainable model due to the incorporation of fuzzy logic into its core. From the definition from \cite{Gunning2019}, an explainable AI (XAI) system should provide capabilities accessible to human understanding, reflecting positively on the health of the system's processes and being able to operate even in unforeseen situations. Although DFS is promoted as an XAI system by definition, this is not the case, as evidenced by works in the literature. This section surveys recent applications of DFS for regression, with the main focus on its structure and if it adheres to the XAI principles.

\subsection{Survey on deep fuzzy systems}
\label{subsec.DeepFuzzyStr}
 
\begin{figure*}[!t]
	\centering
	\includegraphics[width=1\textwidth]{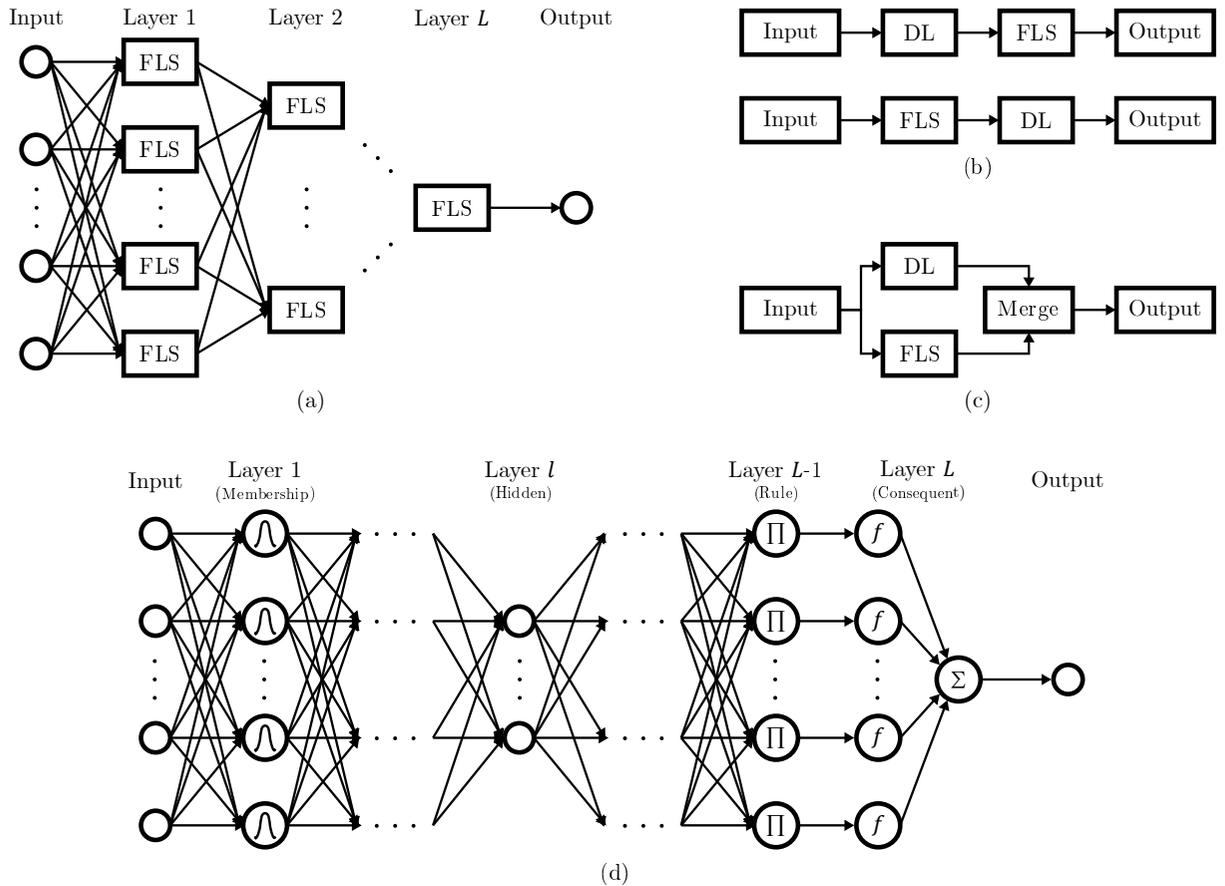}
	\caption{Examples of deep fuzzy system frameworks: (a) multiple FLSs organized hierarchically; (b) sequential ensemble models; (c) parallel ensemble models; (d) fuzzy neural network fully integrated into a deep architecture or mixed with parts of a conventional deep model.}
	\label{fig.dfs}
\end{figure*}

This survey will follow two stages to review the DFS for regression applications. First, the DFS will be categorized according to its structure. Secondly, the models will be categorized according to whether they follow the XAI principles. The structures of deep fuzzy models can be represented in several forms, such as those illustrated in Figure \ref{fig.dfs}. Here, deep fuzzy structures are summarized into two categories: (i) Standard DFS and (ii) Hybrid DFS. A model belongs to the first category when the blocks of fuzzy systems are stacked in series, in parallel, or hierarchically (see Fig. \ref{fig.dfs}a). Also, there are cases where the architecture of such systems resembles neural network architecture, such as the dense DNN architecture (see Fig. \ref{fig.dfs}d). The second category includes hybrid methodologies, where conventional DL models are combined with FLS. The combination of DL and FLS is commonly in an ensemble form (see Fig. \ref{fig.dfs}b,c) or mixed form (see Fig. \ref{fig.dfs}d). 

Aside from the deep fuzzy structures, the discussed works will be investigated whether they are in synergy with the XAI principles. If so, these works will be classified following the categories defined in \cite{Angelov2021b}: understanding and scope of the explainability. In terms of understanding, the models will be classified as (i) transparent or (ii) opaque. Transparent models are those in which the decisions, predictions, or inner functioning are perceptible or are visible; they are considered opaque otherwise. The scope is related to accessing the model's interpretability through post-hoc explanations, classified as (i) local, (ii) global, and (iii) visual. Local explanations facilitate comprehension of small regions of interest in the input space for a given decision/prediction. Global explanations when such desired understanding considers the entire sample space. Visual explanations are required when visual interfaces are needed to demonstrate the influence of features on decisions.

The following sections will discuss the methods containing DFS for regression problems. Section \ref{subsec.DF_MLP} will discuss Standard DFS, while Section \ref{subsec.DF_Var} will discuss Hybrid DFS.

\subsubsection{Standard deep fuzzy systems}
\label{subsec.DF_MLP}

The methods discussed in this section have a DFS structure and were used in any regression problem domain while adhering to the Standard DFS structure. They are discussed in order of structural similarity, with structurally similar methods following each other.

\begin{figure}[!t]
	\centering
	\includegraphics[width=1\columnwidth]{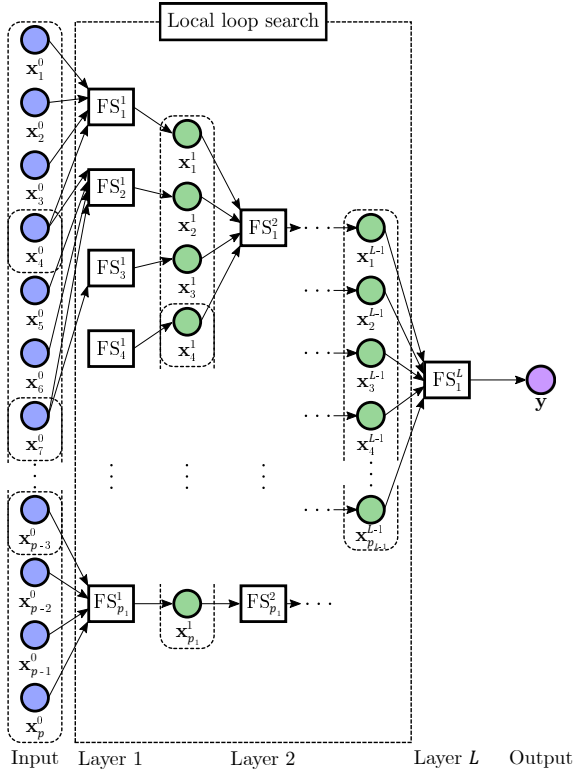}
	\caption{The structure of RLODFS based on grouping and input sharing, adapted from \cite{Huang2021}.}
	\label{fig.rlodfs}
\end{figure}

In \cite{Huang2021}, a model called Randomly Locally Optimized Deep Fuzzy System (RLODFS) is proposed. It is composed of a hierarchical structure of a bottom-up layer-by-layer type with FLS, similar to a fully connected DNN; the structure of RLODFS is shown in Figure \ref{fig.rlodfs}. In RLODFS, the input variables are divided into several fuzzy subsystems, allowing the decrease of the size of the fuzzy rule's antecedent, thus reducing the learning complexity when compared to learning with all input variables at once. The structure of RLODFS shows several groupings with sharing of input variables performed randomly for the fuzzy subsystems of level 1, whose outputs are used as inputs of the fuzzy subsystems of the subsequent levels until reaching the last level fuzzy system (used to estimate the model output). The training of each fuzzy subsystem was followed by the Wang-Mendel algorithm \cite{Wang1992}, which performs the construction of fuzzy rules from a small-scale observation set (data pairs) with input-output mapping. Finally, a random local loop optimization strategy is performed to remove feature combinations and corresponding subsystems with low correlation to achieve fast convergence \cite{Huang2021}. The performance of the RLODFS method was compared with DBN, LSTM, generalized regression neural network (GRNN), among other prediction methods. From the authors' perspective, RLODFS has good interpretability due to its structure, clear physical meaning of its parameters, and the ease of locating fuzzy rules that may fail for future optimizations. However, there is a lack of transparency in using input sharing strategies, which increases the method complexity. Furthermore, the selection of the number of features per fuzzy subsystem needs to be carefully done, whose manual/arbitrary choice may not reflect the real needs of the case study under analysis.

\begin{figure*}[!t]
	\centering
	\includegraphics[width=0.7\textwidth]{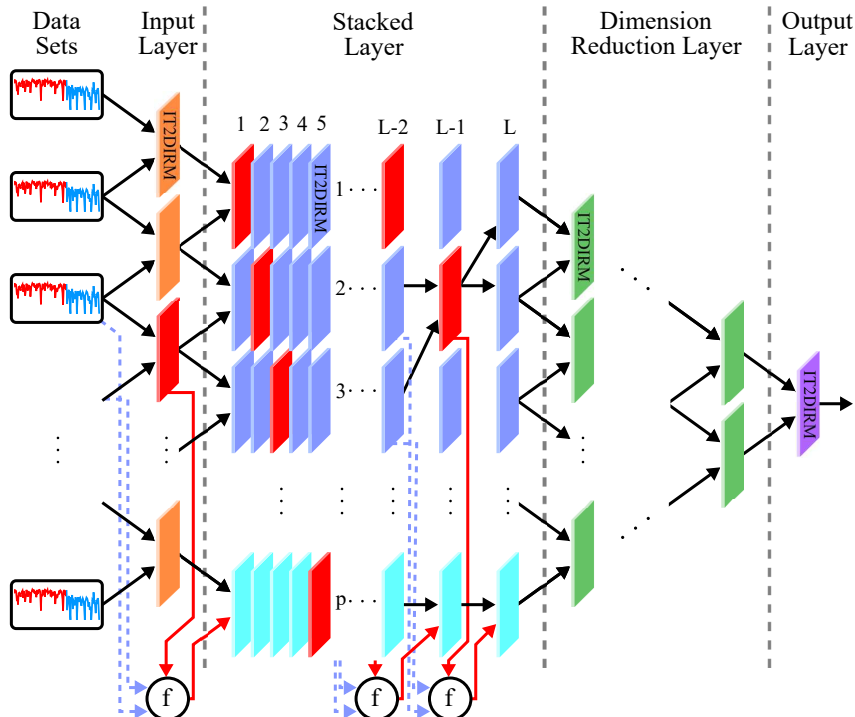}
	\caption{The structure of the double-input rule modules stacked deep interval type-2 fuzzy model (IT2DIRM-DFM), adapted from \cite{Peng2021}.}
	\label{fig.it2dirm_dfm}
\end{figure*}

The study in \cite{Peng2021} proposes a stacked structure composed of double-input rule modules and interval type-2 fuzzy models, abbreviated as IT2DIRM-DFM. The proposed model is illustrated in Figure \ref{fig.it2dirm_dfm}, where four layers are presented: the input layer, which deals directly with the original input data, grouped two-by-two in each rule module; the stacked layer, where the signals coming from the input layer become the inputs of the first layer, whose output becomes the input of the second layer, and so on; the dimension reduction layer, where the width of the IT2DIRM-DFM is hierarchically reduced until it becomes two; and the output layer, where the latest IT2DIRM-DFM model produces the final forecasting results. Still, the authors discuss the interpretability of the model showing layered learning and fuzzy rules composed of only two variables in the antecedent, partitioned into interval type-2 second-order rule partitions. The resulting model was evaluated in two real-world applications, subway passenger and traffic flow, whose results allowed to verify the interpretability from consistent readability of which partitions and bounds of the inputs are used in each fired rule from each rule module. However, the proposed method is only interpretable locally in each rule module and not globally, whose structure does not reflect the depth or number of layers needed to address the experiments. Furthermore, it is not clear the motivations for considering a function, denoted as $\textrm{f}$, in each layer related to the worst-performing module (this is shown at the bottom of Figure \ref{fig.it2dirm_dfm}).

Another work that explores double-input rule modules within a stacked deep fuzzy model (in a hierarchical way) was proposed in \cite{Li2020b}. The authors investigate the interpretability of the resulting model, called DIRM-DFM, with conclusions similar to \cite{Peng2021}, mainly regarding the composition of fuzzy rules. However, in DIRM-DFM, they promote more transparency and simplicity. Some other studies deal with interval type-2 fuzzy models for deep learning in regression problems. In \cite{Qasem2021}, a novel dynamic fractional-order deep learned type-2 FLS was proposed and constructed using singular value decomposition and uncertainty bounds type-reduction. In addition to determining the limit values of the input data (upper and lower singular values), the authors used stability criteria of fractional-order systems, allowing to reduce the necessary number of fuzzy rules and reduce the complexity of nonlinear systems. An evolving recurrent interval type-2 intuitionistic fuzzy neural network (FNN) was proposed in \cite{Luo2019} for time-series prediction. Intuitionistic evaluation, fire strength of membership degree and strategies for adding and removing fuzzy rules were considered to improve uncertainty modeling. Both studies in \cite{Qasem2021} and \cite{Luo2019} did not present an analysis of the interpretability of their models. 

Methods that use multiple neuro-fuzzy systems in a deep hierarchical structure were proposed in \cite{Bodyanskiy2018} and \cite{Bodyanskiy2019}. The work in \cite{Bodyanskiy2018} proposed a deep model that cascades multiple neuro-fuzzy systems modified as multivariable generalized additive models, with application in real ecological time series. The resulting model manages to locally detail the mechanisms of each neuro-fuzzy system in each layer. However, it presents incomplete discussions and results regarding the increase in the network depth and the influence of the inputs and partial outputs of the layers on the final prediction. In \cite{Bodyanskiy2019}, a hybrid cascade neuro-fuzzy network was proposed, which is composed of multiple extended neo-fuzzy neurons with adaptive training designated for online non-stationary data stream handling. Each layer has a generalization node that performs a weighted linear combination to obtain an optimal output signal. The experimental results in electrical loads prediction show the authors' search for better accuracy, although at the cost of increasing membership functions to cover the input space and increasing adjusted parameters (weights).

The work in \cite{Cao2021} proposed a deep learned recurrent type-3 fuzzy system applied for modeling renewable energies (solar panels and wind turbines). The proposed methodology showed good performance compared to other methods, such as multilayer perceptron, type-1 FLS, type-2 FLS, and interval type-3 FLS, despite the lack of a more elaborate discussion of the presented results. Furthermore, transparency is not guaranteed regarding the modeling steps and the influence of various parameters optimized during learning. The authors in \cite{Wang2021} proposed a self-organizing FNN with incremental deep pre-training, abbreviated as IDPT-SOFNN, to promote efficient feature extraction and dynamic adaptation in the structure according to error-reduction rate. In \cite{Wang2020d}, a deep fuzzy cognitive map was proposed for multivariable time-series forecasting, with an analysis of the model's interpretability based on nonlinear and nonmonotonic influences of unknown exogenous factors. The work in \cite{Park2016} proposed a deep FNN composed of an input layer, four hidden layers (membership functions, T-norm operation, linear regression, and aggregation), and an output layer, designed exclusively for intra- and inter-fractional variational prediction for multiple patients' breathing motion.

Table \ref{tab.DFMLPSOTA} summarizes the literature on Standard DFS. They are categorized according to the application domain and regarding its interpretability categories. The survey shows that the application domain is vast, with applications in the domain of industrial systems, power systems, traffic systems, and multivariable benchmark systems.
\begin{table*}[!t]
\centering
\scriptsize
\caption{State-of-the-art on methods with standard deep fuzzy systems for regression problems. XAI: explainable artificial intelligence; Disc.: discussion by the authors (Yes/No); Und.: how understandable is the model, whether it is transparent (T) or opaque (O); Scope: in post-hoc scope, if the model promotes local explanations (L), global explanations (G) or visual explanations (V).}
{\renewcommand{\arraystretch}{1.5}
\begin{tabularx}{\textwidth}{lXXlll}
\hline
\multirow{2}{*}{Reference} & \multirow{2}{*}{Approach} & \multirow{2}{*}{Problem} & \multicolumn{3}{l}{XAI} \\ \cline{4-6}  &  &  & \multicolumn{1}{l}{Disc.} & \multicolumn{1}{l}{Und.} & Scope
\\ \hline
\cite{Huang2021} & 
    Hierarchical Randomly Locally Optimized Deep Fuzzy System with input sharing
    &  
    Prediction of 12 real-world datasets
    &
    Yes
    & 
    T
    &
    L
\\
\cite{Peng2021} & 
    Double-input rule modules stacked deep interval type-2 fuzzy model
    &  
    Time-series forecasting
    &
    Yes
    & 
    T
    &
    L
\\
\cite{Li2020b} & 
    Double-input-rule-modules stacked deep fuzzy model
    &  
    Prediction of short-term photovoltaic power generation
    &
    Yes
    & 
    T
    &
    L,G
\\
\cite{Qasem2021} & 
    Dynamic fractional-order deep learned type-2 fuzzy logic system
    &  
    Benchmark system prediction
    &
    No
    & 
    O
    &
    -
\\
\cite{Luo2019} & 
    Evolving recurrent interval type-2 intuitionistic FNN
    &  
    Time-series forecasting
    &
    No
    & 
    T
    &
    -
\\
\cite{Bodyanskiy2018} & 
    Deep Stacking Convex Neuro-Fuzzy System
    &  
    Real ecological time-series forecasting
    &
    No
    & 
    T
    &
    L
\\
\cite{Bodyanskiy2019} & 
    Hybrid cascade neuro-fuzzy scheme (ensemble of extended neo-fuzzy neurons)
    &  
    Time-series forecasting (electrical loads)
    &
    No
    & 
    T
    &
    -
\\
\cite{Cao2021} & 
    Deep learned recurrent type-3 fuzzy system
    &  
    Renewable energy modeling (solar panels and wind turbines)
    &
    No
    & 
    O
    &
    -
\\
\cite{Wang2021} & 
    Self-Organizing Deep FNN
    &  
    Nonlinear system modeling
    &
    No
    & 
    T
    &
    V
\\
\cite{Wang2020d} & 
    Deep Fuzzy cognition map model (with an alternate function gradient descent algorithm)
    &  
    Multivariate time-series forecasting
    &
    Yes
    & 
    T
    &
    V
\\
\cite{Park2016} & 
    Fuzzy Deep Learning architecture (with four hidden layers)
    &  
    Intra- and Inter-fractional variation prediction of Lung Tumors
    &
    No
    & 
    T
    &
    L,V
\\ \hline
\end{tabularx}
}
\label{tab.DFMLPSOTA}
\end{table*}
Table \ref{tab.DFMLPSOTA} shows that only four out of eleven works discuss interpretability, despite most of their proposed methods being transparent. Also, not all methods provide post-hoc explanations, and of these, the local scope is more frequent. The Standard DFS architecture is easy to implement, with flexibility in the construction of fuzzy rules, having a similar structure to feedforward neural networks. However, these methods may suffer from loss of interpretability when using a non-intuitive hierarchical structure, the lack of investigation of interconnections between the variables involved, and the limitation of fuzzy rules that can disturb the estimation by not covering all operating regions (coverage of input data space) \cite{Magdalena2019}.

\subsubsection{Hybrid deep fuzzy systems}
\label{subsec.DF_Var}

The Hybrid DFSs are discussed in this section. The selection and order of works follow the same principles used for Standard DFS. The common DL architectures used in combination with fuzzy systems are the Deep Belief Networks, Autoencoders, Long Short-Term Memory networks, and Echo State Networks.

\begin{figure*}[!t]
	\centering
	\includegraphics[width=1\textwidth]{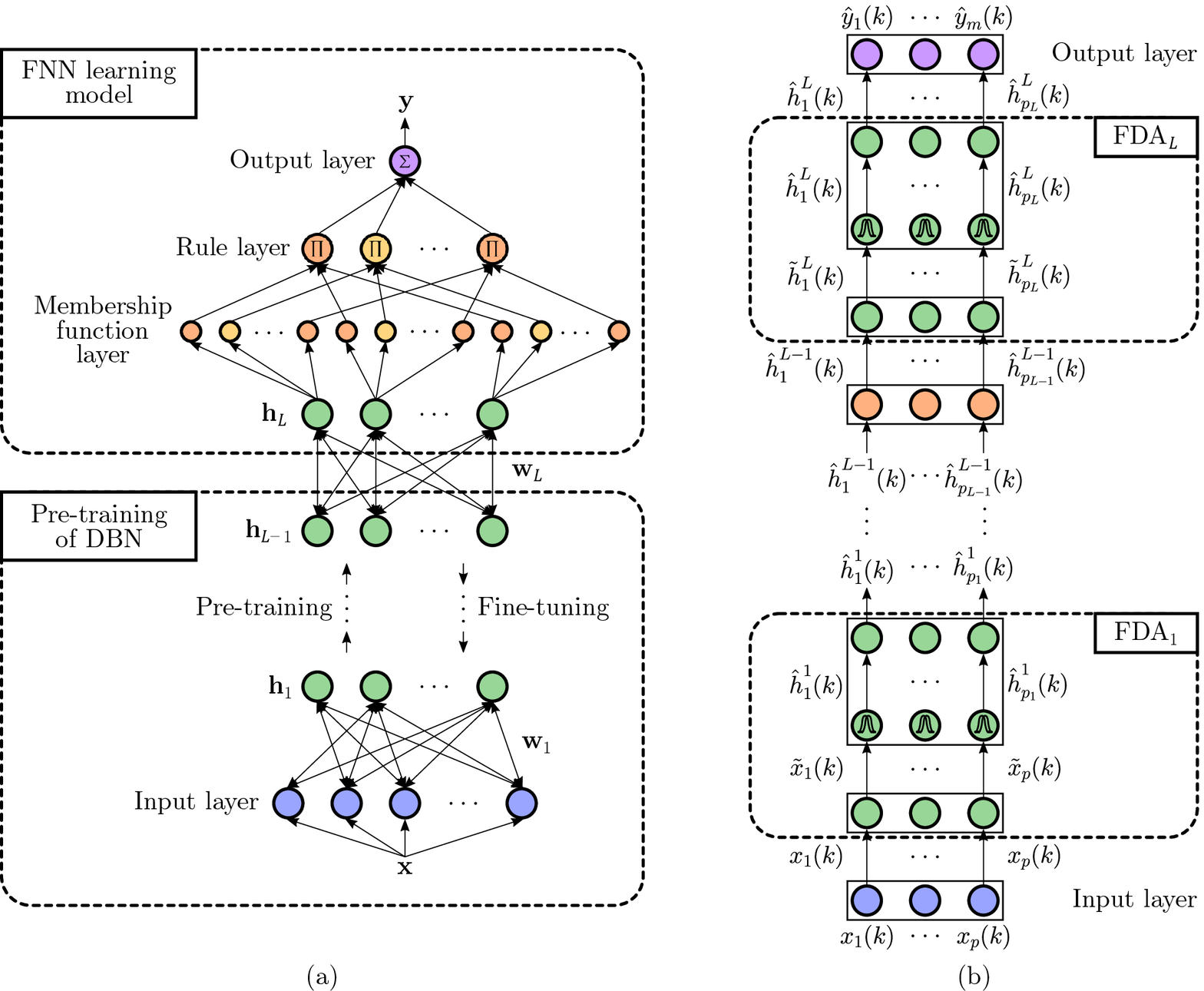}
	\caption{Examples of FLS application with conventional deep models: (a) with deep belief networks, adapted from \cite{Wang2020}; and (b) with denoising autoencoders, adapted from \cite{Han2020}.}
	\label{fig.deepfuzzyvariants}
\end{figure*}

In \cite{Wang2020}, it is proposed a sparse Deep Belief Network (SDBN) with FNN for nonlinear system modeling. The SDBN is considered for unsupervised learning and pre-training to perform fast weight-initialization and improve modeling robustness. The FNN is used as supervised learning to reduce layer-by-layer complexity. As shown in Figure \ref{fig.deepfuzzyvariants}, the structure of the DBN resembles the structure shown in Figure \ref{fig.dbn}, except for considering additional constraints (sparsity) used to penalize fluctuations of values along the hidden neurons. The proposed method performed better compared to other similar methods, such as transfer learning-based growing DBN \cite{Wang2019}, DBN-based echo-state network \cite{Sun2017}, and self-organizing cascade neural network \cite{Li2016}. However, the authors noticed various fluctuations in the assignment of hyperparameters, which can compromise the stability of the proposed model, making it necessary to dynamically and robustly improve its structure to these fluctuations. In terms of interpretability, the proposed model guarantees a moderate number of membership functions and rules, allowing good consistency and readability of what happens within the FNN structure. The same cannot be said for the DBN structure, which is not intuitive in sparse representation to decide which features are more valuable than others.

The authors in \cite{Han2020} proposed a novel robust deep neural network (RDNN) for regression problems involving nonlinear systems, with the implementation of three strategies: a fuzzy denoising autoencoder (FDA) as a base-building unit for RDNN, improving the ability to represent uncertainties; a compact parameter strategy (CPS), designed to reconstruct the parameters of the FDA, reducing unnecessary learning parameters; and an adaptive backpropagation (ABP) algorithm to update RDNN parameters with fast convergence. As shown in Figure \ref{fig.deepfuzzyvariants}, FDA has an input layer, a fuzzy layer (built by Gaussian membership functions), and an output layer, which can be divided into an encoder (``input'' to ``fuzzy'') and a decoder (``fuzzy'' to ``output''). Furthermore, at each FDA, due to the nature of this autoencoder, the data is partially corrupted by noises (represented with a ``tilde'' symbol) and is reconstructed using CPS (represented with a ``hat'' symbol) with associated parameters (\eg, encoder/decoder weights and biases) are adjusted via ABP. Regarding the interpretability of the proposed model, there was no appropriate discussion by the authors, as they focused mainly on model performance/accuracy in the presence of uncertainties. In addition to a fuzzy rule base not being defined, the number of fuzzy neurons in each FDA is manually initialized and adapts through redundancies during the reconstruction of parameters via CPS. Finally, the architecture of the proposed model is not intuitive based on the mechanisms related to the representation ability of each FDA, which would be crucial to determine the influence of input variables and learning parameters on the outputs.

Some hybrid methods using FLS and conventional deep models have been applied in the literature for traffic flow prediction. In \cite{George2020}, the authors developed an algorithm based on Dolphin Echolocation optimization \cite{Borkar2016}, where the input features are fuzzified into membership functions to obtain chronological data, whose integration goes into the weight update process of the proposed algorithm converging to a globally optimal solution with a Deep Belief Network. The study in \cite{Chen2019b} combined fuzzy information granulation and a deep neural network to represent the temporal-spatial correlation of mass traffic data and be able to adapt to noisy data. A Stacked Autoencoder is used to obtain the prediction results based on processed granules that have a good capacity for interpretation, which have not been discussed by the authors.

Other methods were implemented for energy forecasting, with the usual application of LSTM networks. A novel fuzzy seasonal LSTM was proposed in \cite{Liao2021}, where the fuzzy seasonality index \cite{Chang1997} and a decomposition method were employed to solve the seasonal time-series problem in a monthly wind power output dataset. In \cite{Imani2021}, the authors used an LSTM network with rough set theory \cite{Pawlak1982} and interval type-2 fuzzy sets for short-term wind speed forecasting, with the aid of mutual information approach for efficient variable input selection. A novel ultra-short-term photovoltaic power forecasting method was proposed in \cite{Liu2021d}, where a T-S fuzzy model is developed using a fuzzy c-means clustering algorithm and deep belief networks. The studies in \cite{Liao2021}, \cite{Imani2021} and \cite{Liu2021d} did not discuss the interpretability of their models, which have a structure with an ensemble characteristic whose fuzzy part has an affinity for good coverage of the input space and good capture of data uncertainty.

The work in \cite{Chimatapu2020} proposed a deep type-2 FLS (D2FLS) architecture with greedy layer-wise training for high dimensional input data. The authors showed how to extract interpretable explanations related to the contribution of fuzzy rules to the final prediction developed for a two-layer D2FLS. However, the authors opted for a large number of fuzzy rules (100, in this case) that impair interpretability by increasing the complexity of the model. The authors in \cite{SN2018} proposed an ensemble model composed of an Echo State Network (ESN), a T-S fuzzy model, and differential evolution for time-series forecasting problems. The differential evolution method, used to optimize the weight coefficients of the model, managed to reduce the number of fuzzy rules. The interpretability of the model can be impaired due to the structure with an ensemble characteristic, which does not provide enough transparency for learning. A method based on the T-S fuzzy model and ESN for the identification of nonlinear systems was developed in \cite{Mahmoud2022}. The authors chose to balance model complexity and performance based on the parameters involved for learning (\eg, number of fuzzy rules and reservoir size), which resulted in better results in comparison with other methods (\eg, traditional ESN and hybrid fuzzy ESN), despite the lack of discussion on interpretability.

In \cite{Wu2020}, a framework based on a sparse autoencoder (SpAE) and a high-order fuzzy cognitive map (HFCM) is proposed for time-series forecasting problems. SpAE is used to extract features from the original data, and these features are via HFCM. Another study that uses SpAE with FLS was proposed in \cite{Sevakula2015} and implemented in time-series forecasting and classification problems. The proposed method uses a method to reduce fuzzy rules by reducing the data dimensionality with SpAE. Both studies in \cite{Wu2020} and \cite{Sevakula2015} do not consider addressing the interpretability of the proposed models, which present good directions in data partitioning and construction of fuzzy rules but are not intuitive in feature extraction with sparse representation.

Table \ref{tab.DFVARSOTA} summarizes the works presented in this section, in addition to evaluating them within the context of explainable artificial intelligence (XAI) systems.
\begin{table*}[!t]
\centering
\scriptsize
\caption{State-of-the-art on hybrid methods using fuzzy logic systems and conventional deep models for regression problems. XAI: explainable artificial intelligence; Disc.: discussion by the authors (Yes/No); Und.: how understandable is the model, whether it is transparent (T) or opaque (O); Scope: in post-hoc scope, if the model promotes local explanations (L), global explanations (G) or visual explanations (V).}
{\renewcommand{\arraystretch}{1.5}%{1.2}
\begin{tabularx}{\textwidth}{lXXlll}
\hline
\multirow{2}{*}{Reference} & \multirow{2}{*}{Approach} & \multirow{2}{*}{Problem} & \multicolumn{3}{l}{XAI} \\ \cline{4-6}  &  &  & \multicolumn{1}{l}{Disc.} & \multicolumn{1}{l}{Und.} & Scope 
\\ \hline
\cite{Wang2020} & 
    Sparse Deep Belief Network with FNN
    &  
    Nonlinear system modeling
    &
    No
    &
    T
    &
    -
\\
\cite{Han2020} & 
    Fuzzy denoising autoencoder with a compact parameter strategy and an adaptive backpropagation algorithm
    &  
    Prediction of multi-input single-output systems
    &
    No
    &
    O
    &
    -
\\
\cite{George2020} & 
    Chronological Dolphin Echolocation-Fuzzy and Deep Belief Network
    &  
    Traffic flow prediction in intelligent transportation system
    &
    No
    &
    O
    &
    -
\\
\cite{Chen2019b} & 
    Fuzzy information granulation with Stacked Autoencoder
    &  
    Traffic flow prediction
    &
    No
    &
    T
    &
    -
\\
\cite{Liao2021} & 
    Fuzzy seasonal long short-term memory network
    &  
    Wind power forecasting
    &
    No
    &
    O
    &
    -
\\
\cite{Imani2021} & 
    Long Short-Term Memory model hybridized with rough and fuzzy set theory
    &  
    Wind speed forecasting
    &
    No
    &
    O
    &
    -
\\
\cite{Liu2021d} & 
    T-S fuzzy model with fuzzy c-means and Deep Belief Network
    &  
    Forecasting of photovoltaic power generation
    &
    No
    &
    T
    &
    -
\\
\cite{Chimatapu2020} & 
    Deep Type-2 Fuzzy Logic System, trained as Stacked Autoencoder
    &  
    Benchmark system prediction and classification
    &
    Yes
    &
    T
    &
    G,V
\\
\cite{SN2018} & 
    Differential Evolution based Fuzzy Echo State Network
    &  
    Time-series forecasting
    &
    No
    &
    O
    &
    -
\\
\cite{Mahmoud2022} & 
    T-S fuzzy model with Echo State Network
    &  
    Benchmark system prediction
    &
    No
    &
    T
    &
    -
\\
\cite{Wu2020} & 
    Fuzzy Cognitive Maps with Sparse Autoencoders
    &  
    Time-series forecasting
    &
    No
    &
    T
    &
    -
\\
\cite{Sevakula2015} & 
    Fuzzy rule reduction with Stacked Sparse Autoencoders
    &  
    Time-series prediction and classification
    &
    No
    &
    O
    &
    -
\\ \hline
\end{tabularx}
}
\label{tab.DFVARSOTA}
\end{table*}\relax
Only two works discuss the interpretability in Hybrid DFS, half of the discussed works are opaque, and only one presents post-hoc explanations. This fact occurs since the combination with DNN brings a new layer of black-box to the system. However, the Hybrid DFS methods categorized as transparent show that the fuzzy component can promote good interpretability with efficient input-output mappings and the construction of rules to cover the universe of discourse for a given system. Inherently interpretable methodologies are difficult to achieve only with an ensemble of multiple methods, as seen in recent studies, in addition to the challenges of reducing the time complexity of systems, which is slightly mitigated by the reduction of fuzzy rules \cite{Ojha2019}.

\section{Conclusion}
\label{sec.Conclusions}

This study surveyed the literature on deep fuzzy systems (DFSs) for regression applications within the context of explainable artificial intelligence (XAI) systems (with an emphasis on interpretability). For the survey, the DFSs were categorized as (i) Standard DFS and (ii) Hybrid DFS. Regarding the interpretability, each method was categorized as to whether it is transparent or not and whether it has post-hoc explanations (under the definition of \cite{Angelov2021b}). Standard DFSs have been shown to promote more interpretability of their models when compared to Hybrid DFSs, according to the survey. Indeed, Standard DFSs are based on fundamental fuzzy logic systems, which are inherently interpretable, whereas Hybrid DFSs include conventional deep learning (DL) methods, which lack flexibility in promoting interpretability. Furthermore, the DFS is frequently referred to as interpretable by default, but only 5 of the 23 works surveyed here actually addressed this issue. The remaining works had a common goal: to improve prediction accuracy using their proposed methods. However, this survey presented the potential of using Standard DFS as a base for developing accurate models while promoting interpretability since hybrid models are not straightforward.

\backmatter

\bmhead{Acknowledgments}

This work has been supported by Funda{\c c}{\~a}o para a Ci{\^e}ncia e a Tecnologia (FCT) under the project UIDB/00048/2020. Jorge S. S. J{\'u}nior is supported by Funda{\c c}{\~a}o para a Ci{\^e}ncia e a Tecnologia (FCT) under the grant ref. 2021.04917.BD.

\section*{Competing Interests}

The authors declare no competing Interests.

\bibliography{Ref_SurveyDFS_JorgeSSJunior}

\end{document}